\documentclass[10pt,logo,copyright]{giga-report}
\usepackage[authoryear,sort&compress,round]{natbib}

\usepackage[utf8]{inputenc} %
\usepackage[T1]{fontenc}    %

\usepackage{parskip}        %
\usepackage{url}            %
\usepackage{booktabs}       %
\usepackage{amsfonts}       %
\usepackage{nicefrac}       %
\usepackage{microtype}      %
\usepackage{xcolor}         %
\usepackage[dvipsnames]{xcolor} %
\usepackage{graphicx}
\usepackage{animate}        %
\usepackage{subcaption}
\usepackage{tabularx}
\usepackage{makecell}
\usepackage{adjustbox}
\usepackage{setspace}
\usepackage{todonotes}
\usepackage{colortbl}
\usepackage{wrapfig}

\newcolumntype{M}[1]{>{\centering\arraybackslash}m{#1}}
\usepackage{float}
\usepackage{tikz}
\usetikzlibrary{positioning,shapes,arrows}
\usepackage{amsmath,amsfonts,bm, bbm,leftindex}
\usepackage{multirow}
\usepackage{comment}
\usepackage{gensymb}
\usepackage{lipsum}
\usetikzlibrary{arrows.meta, positioning, fit}
\usepackage[para]{threeparttable}
\usepackage{tikz}
\usetikzlibrary{tikzmark}

\def\eqref#1{equation~\ref{#1}}

\def\1{\bm{1}}

\DeclareMathAlphabet{\mathsfit}{\encodingdefault}{\sfdefault}{m}{sl}
\SetMathAlphabet{\mathsfit}{bold}{\encodingdefault}{\sfdefault}{bx}{n}

\makeatletter
\let\save@mathaccent\mathaccent
\newcommand*\if@single[3]{%
  \setbox0\hbox{${\mathaccent"0362{#1}}^H$}%
  \setbox2\hbox{${\mathaccent"0362{\kern0pt#1}}^H$}%
  \ifdim\ht0=\ht2 #3\else #2\fi
  }
\newcommand*\rel@kern[1]{\kern#1\dimexpr\macc@kerna}
\newcommand*\widebar[1]{\@ifnextchar^{{\wide@bar{#1}{0}}}{\wide@bar{#1}{1}}}
\newcommand*\wide@bar[2]{\if@single{#1}{\wide@bar@{#1}{#2}{1}}{\wide@bar@{#1}{#2}{2}}}
\newcommand*\wide@bar@[3]{%
  \begingroup
  \def\mathaccent##1##2{%
    \let\mathaccent\save@mathaccent
    \if#32 \let\macc@nucleus\first@char \fi
    \setbox\z@\hbox{$\macc@style{\macc@nucleus}_{}$}%
    \setbox\tw@\hbox{$\macc@style{\macc@nucleus}{}_{}$}%
    \dimen@\wd\tw@
    \advance\dimen@-\wd\z@
    \divide\dimen@ 3
    \@tempdima\wd\tw@
    \advance\@tempdima-\scriptspace
    \divide\@tempdima 10
    \advance\dimen@-\@tempdima
    \ifdim\dimen@>\z@ \dimen@0pt\fi
    \rel@kern{0.6}\kern-\dimen@
    \if#31
      \overline{\rel@kern{-0.6}\kern\dimen@\macc@nucleus\rel@kern{0.4}\kern\dimen@}%
      \advance\dimen@0.4\dimexpr\macc@kerna
      \let\final@kern#2%
      \ifdim\dimen@<\z@ \let\final@kern1\fi
      \if\final@kern1 \kern-\dimen@\fi
    \else
      \overline{\rel@kern{-0.6}\kern\dimen@#1}%
    \fi
  }%
  \macc@depth\@ne
  \let\math@bgroup\@empty \let\math@egroup\macc@set@skewchar
  \mathsurround\z@ \frozen@everymath{\mathgroup\macc@group\relax}%
  \macc@set@skewchar\relax
  \let\mathaccentV\macc@nested@a
  \if#31
    \macc@nested@a\relax111{#1}%
  \else
    \def\gobble@till@marker##1\endmarker{}%
    \futurelet\first@char\gobble@till@marker#1\endmarker
    \ifcat\noexpand\first@char A\else
      \def\first@char{}%
    \fi
    \macc@nested@a\relax111{\first@char}%
  \fi
  \endgroup
}
\makeatother

\definecolor{darkred}{rgb}{0.7, 0.0, 0.0}
\newcommand{\cmark}{\textcolor[HTML]{9C27B0}{\ding{51}}}

\usepackage{amsmath} 
\usepackage{amssymb}
\usepackage{dsfont}
\usepackage{pifont}

\usepackage[nameinlink]{cleveref}
\crefname{equation}{Eq.}{Eqs.}
\crefname{figure}{Fig.}{Figs.}
\crefname{section}{Sec.}{Sec.}
\crefname{appendix}{App.}{App.}
\crefname{table}{Tab.}{Tabs.}
\crefname{algorithm}{Algo}{Algo}
\crefname{thm}{Thm}{Thm}
\Crefname{thm}{Thm}{Thm}
\crefname{prop}{Prop}{Prop}

\newcommand{\crefnames}[3]{%
  \@for\next:=#1\do{%
    \expandafter\crefname\expandafter{\next}{#2}{#3}%
  }%
}

\title{GigaWorld-0: World Models as Data Engine to Empower Embodied AI}

\author{
\centerline{GigaAI} 
\centerline{{Project Page: \href{https://giga-world-0.github.io}{https://giga-world-0.github.io}}} 
\footnotesize
\textbf{GigaWorld Team (alphabetical order)}:
\normalfont
Angen Ye,  
Boyuan Wang,  
Chaojun Ni,  
Guan Huang,  
Guosheng Zhao,  
Haoyun Li,  
Jiagang Zhu,  
Kerui Li,
Mengyuan Xu,
Qiuping Deng,  
Siting Wang,
Wenkang Qin,  
Xinze Chen,  
Xiaofeng Wang,  
Yankai Wang,
Yu Cao,  
Yifan Chang,
Yuan Xu,  
Yun Ye,  
Yang Wang,
Yukun Zhou,  
Zhengyuan Zhang,
Zhehao Dong,  
Zheng Zhu.
\vspace{-2em}
}

\begin{document}
\maketitle

\begin{center}
    \
    \centering
    \captionsetup{type=figure, justification=justified, singlelinecheck=false}
    \includegraphics[width=1\linewidth]{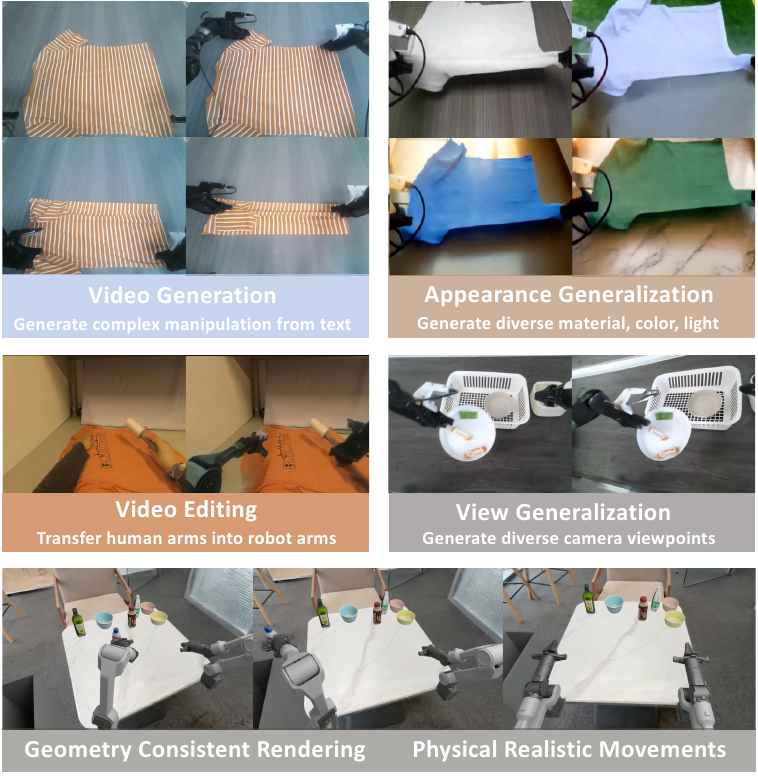}
    \caption{An overview of \textit{GigaWorld-0} applications. Using video generation, it dynamically alters appearance and viewpoints. \textit{GigaWorld-0} also facilitates converting human demonstration videos into robotic manipulation trajectories. Furthermore, through 3D scene generation and reconstruction, \textit{GigaWorld-0} supports physically realistic motion planning and produces geometrically consistent renderings to empower embodied AI.}
    \label{fig:teaser}
\end{center}

\newpage
\begin{abstract}
World models are emerging as a foundational paradigm for scalable, data-efficient embodied AI. In this work, we present \textit{GigaWorld-0}, a unified world model framework designed explicitly as a \textit{data engine} for Vision-Language-Action (VLA) learning. \textit{GigaWorld-0} integrates two synergistic components: \textit{GigaWorld-0-Video}, which leverages large-scale video generation to produce diverse, texture-rich, and temporally coherent embodied sequences under fine-grained control of appearance, camera viewpoint, and action semantics; and \textit{GigaWorld-0-3D}, which combines 3D generative modeling, 3D Gaussian Splatting reconstruction, physically differentiable system identification, and executable motion planning to ensure geometric consistency and physical realism. Their joint optimization enables the scalable synthesis of embodied interaction data that is visually compelling, spatially coherent, physically plausible, and instruction-aligned. Training at scale is made feasible through our efficient \textit{GigaTrain} framework, which exploits FP8-precision and sparse attention to drastically reduce memory and compute requirements. 
We conduct comprehensive evaluations showing that \textit{GigaWorld-0} generates high-quality, diverse, and controllable data across multiple dimensions. Critically, VLA model (e.g., {GigaBrain-0}) trained on \textit{GigaWorld-0}-generated data achieve strong real-world performance, significantly improving generalization and task success on physical robots without any real-world interaction during training.
\end{abstract}

\abscontent
\section{Introduction}

World models have emerged as a foundational component for advancing embodied AI~\citep{cosmospredict,issora}, functioning as high-fidelity simulators that bridge the gap between synthetic and real-world environments. By modeling the dynamics, appearance, and spatial structure of physical scenes, world models enable the efficient, scalable, and cost-effective generation of high-quality, diverse data under a broad spectrum of controllable conditions. This capability significantly alleviates the data bottleneck traditionally faced by embodied agents, which rely heavily on expensive real-world data collection. In this work, we present \textit{GigaWorld-0} and underscore the role of the world model as a powerful data engine, a scalable, controllable, and photorealistic training data source that significantly empowers the learning of embodied AI.

Accurate modeling of texture, geometry, and physical dynamics is fundamental to the fidelity and utility of any world model aiming to serve as a high-quality proxy for the physical world. To address these requirements in a scalable and controllable manner, the introduced \textit{GigaWorld-0} comprises \textit{GigaWorld-0-Video} and \textit{GigaWorld-0-3D}.
\textit{GigaWorld-0-Video} leverages video generation models to synthesize temporally coherent, photorealistic visual sequences with rich texture detail, diverse scene content, and fine-grained control over appearance (e.g., texture, material, light), object placement and camera viewpoints. This enables the efficient generation of large-scale, high-quality 2D observation data under a broad distribution of real-world conditions.
In parallel, \textit{GigaWorld-0-3D} explicitly enforces 3D geometric consistency and physical plausibility by integrating 3D reconstruction models and physics-aware simulation priors. It ensures spatial coherence across viewpoints, models object rigidity and deformability, and respects physical constraints such as collision, gravity, and contact dynamics. The coupling of \textit{GigaWorld-0-Video} and \textit{GigaWorld-0-3D} further yields a unified data generation pipeline capable of producing embodied interaction data that is simultaneously texture-rich, geometrically consistent, physically grounded, and dynamically realistic. 

Specifically, the model suite of \textit{GigaWorld-0} is summarized in Tab.~\ref{tab:GigaWorld-0_models}. The \textit{GigaWorld-0-Video} comprises four generative models: \textit{GigaWorld-0-Video-Dreamer}, \textit{GigaWorld-0-Video-AppearanceTransfer}, \textit{GigaWorld-0-Video-ViewTransfer}, and \textit{GigaWorld-0-Video-MimicTransfer}. Among them, \textit{GigaWorld-0-Video-Dreamer} serves as our Video Foundation Model—a Mixture-of-Experts (MoE) architecture for image-text-to-video (IT2V) generation, trained on a large-scale corpus of embodied interaction data. Building upon this foundation, the three post-training adaptation models, \textit{GigaWorld-0-Video-AppearanceTransfer}, \textit{GigaWorld-0-Video-ViewTransfer}, and \textit{GigaWorld-0-Video-MimicTransfer}, each incorporate dedicated controllable branches to enable generalization across appearance (e.g., texture and lighting), camera viewpoints, and action modalities, respectively. Notably, \textit{GigaWorld-0-Video-MimicTransfer} translates first-person human manipulation demonstrations into robot-executable trajectories, facilitating cross-embodiment generalization. To better suit embodied manipulation scenarios, we further extend the training pipeline with multi-view video generation, FP8-precision training acceleration, denoising-step distillation and FP8-efficient inference.
Complementing the video stream, \textit{GigaWorld-0-3D} constructs physically grounded 3D scenes through a modular pipeline: \textit{GigaWorld-0-3D-FG} generates foreground assets via a 3D generative model. \textit{GigaWorld-0-3D-BG} reconstructs background environments using 3D Gaussian Splatting (3DGS). \textit{GigaWorld-0-3D-Phys} models the physical properties of interactable objects and performs differentiable system identification for the robotic arm. \textit{GigaWorld-0-3D-Act} calculates arm motions to produce complete, executable manipulation sequences.

\begin{table}[t]
\centering
\small
\caption{Components of \textit{GigaWorld-0} and their functions.}
\label{tab:GigaWorld-0_models}
\begin{tabular}{@{}ll@{}}
\toprule
\textbf{Model Name} & \textbf{Function} \\
\midrule

\textit{GigaWorld-0-Video-Dreamer} & 
Image-text-to-video foundation model for embodied scenes. \\
\textit{GigaWorld-0-Video-AppearanceTransfer} & 
Text-guided appearance transfer, edits texture, material, lighting. \\
\textit{GigaWorld-0-Video-ViewTransfer} & 
Renders videos from user-specified camera extrinsics. \\
\textit{GigaWorld-0-Video-MimicTransfer} & 
Translates egocentric human demonstration to robot arm trajectories. \\
\midrule

\textit{GigaWorld-0-3D-FG} & 
Generates 3D assets of foreground manipulable objects. \\
\textit{GigaWorld-0-3D-BG} & 
Reconstructs backgrounds via 3D Gaussian Splatting (3DGS). \\
\textit{GigaWorld-0-3D-Phys} & 
Models object physics and performs differentiable system identification. \\
\textit{GigaWorld-0-3D-Act} & 
Synthesizes executable, physically consistent arm motions. \\
\bottomrule
\end{tabular}
\end{table}

The large-scale training of \textit{GigaWorld-0-Video} is efficiently enabled by \textit{GigaTrain}, our training framework leveraging FP8-precision and sparse attention to accelerate the training process.
We conduct extensive experiments to validate the effectiveness of \textit{GigaWorld-0} from multiple dimensions, including physical plausibility, geometric consistency, text-to-video alignment, multi-view coherence, and visual fidelity. Quantitative and qualitative results demonstrate that \textit{GigaWorld-0} achieves state-of-the-art performance across these metrics, setting a new benchmark for synthetic data generation in embodied settings. Moreover, we evaluate policies trained on \textit{GigaWorld-0}-generated data in real-world robotic environments, confirming that such synthetic data significantly enhances the training of Vision-Language-Action (VLA) models, leading to improved task success rates, robustness, and generalization under diverse conditions. 

The proposed \textit{GigaWorld-0} as a powerful data engine for embodied AI. To foster community progress, we will open-source the models and data generation pipeline. We believe world models remain a vast and underexplored frontier; promising directions such as World Models as Policy Environments and World Models for Policy Generation warrant collaborative exploration by the broader research community.
\section{Related Work}

Recent progress in world model research~\citep{hunyuanvideo,wan,cosmos,cosmospredict,vjepa2,genieEnvisioner,enerverseac,dreamgen} has accelerated the adoption of generated data as a training source to facilitate the learning of embodied AI~\citep{issora}. In autonomous driving, generative world models are now routinely used to create complex, safety-critical traffic scenarios; representative frameworks include~\citep{drivedreamer,drivedreamer2,gaia,gaia2,magicdrive,vista,drivedreamer4d,recondreamer,recondreamer++,cosmosdrivedream,recondreamerrl}. Similarly, in robotics, where real-world data collection is constrained by hardware availability, safety, and labor costs, generative approaches offer a scalable alternative. Several works~\citep{unisim,unipi,robodreamer,vidar,dreamgen} utilize natural language instructions to forecast plausible future observations, subsequently deriving low-level motor commands through inverse dynamics or action decoding. To enhance geometric and temporal fidelity, methods such as TesserAct~\citep{tesseract} and Robot4DGen~\citep{Robot4DGen} propose unified multimodal generation pipelines that jointly synthesize aligned RGB, depth, surface normals, and 3D point clouds, enabling coherent 4D scene reconstructions that substantially improve policy learning over RGB-only training. Additional gains in environmental diversity are achieved via background inpainting techniques~\citep{rebot,roboengine}, which modify scene textures, and video-to-video translation frameworks~\citep{robotransfer,embodiedreamer,cosmos,mimicdreamer,egodemogen,emma} that adapt visual styles or dynamics across domains. Building on this trend, \textit{GigaWorld-0} harnesses world models to generate highly diverse synthetic data spanning variations in texture, material, lighting, object layout, and camera pose, thereby offering a rich and generalizable training signal for VLA learning.
\section{GigaWorld-0 Models}
This section first introduces the \textit{GigaWorld-0-Video} series, followed by the \textit{GigaWorld-0-3D} series. \textit{GigaWorld-0-Video} leverages video generation models to synthesize photorealistic sequences with control over appearance, object placement, and camera viewpoints, enabling large-scale, high-quality data generation under diverse real-world conditions. In parallel, \textit{GigaWorld-0-3D} enforces 3D geometric consistency and physical plausibility via 3D representation, ensuring spatial coherence, modeling object rigidity/deformability, and respecting physical constraints like contact dynamics.

\subsection{GigaWorld-0-Video}
Foundation models for video generation in embodied scenarios must exhibit a profound understanding of such environments and be capable of efficiently synthesizing plausible embodied interaction videos conditioned on diverse control signals. In contrast to existing video generation models~\citep{wan,hunyuanvideo,cogvideox,opensora,cosmospredict}, which primarily scale up model parameters, the \textit{GigaWorld-0-Video} series achieve lower training costs and reduced inference latency through a combination of sparse attention mechanisms, a mixture-of-experts (MoE) architecture, FP8-precision training and inference, and diffusion step distillation.

\begin{figure}[!t]
    \centering
    \includegraphics[width=1\linewidth]{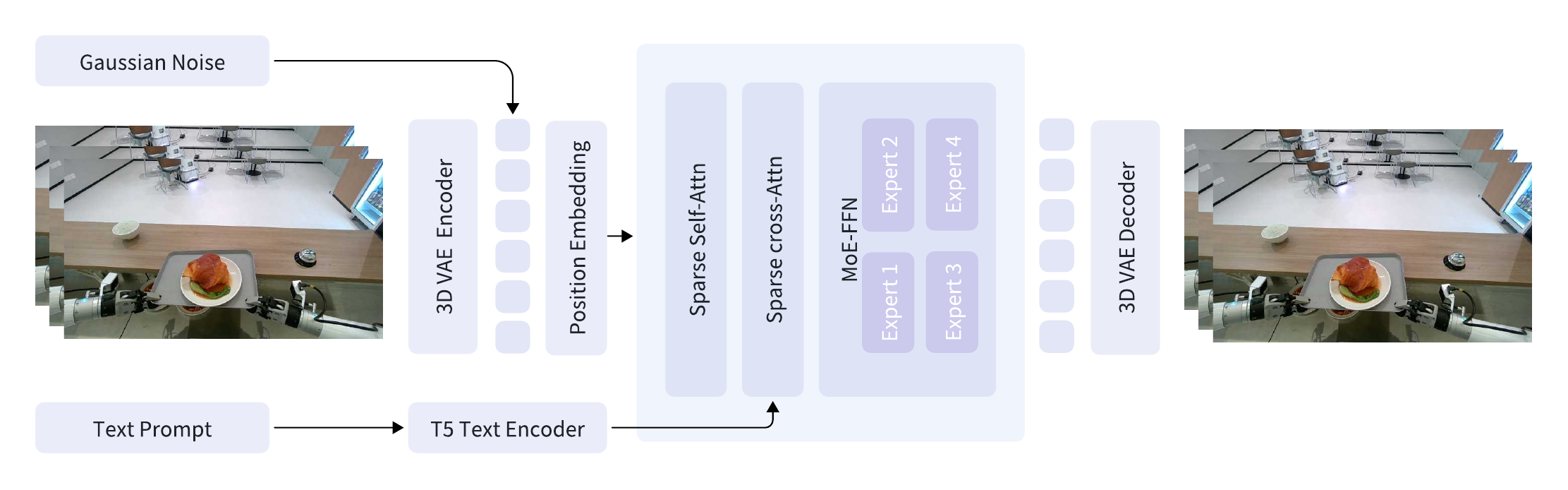}
    \captionsetup{type=figure, justification=justified, singlelinecheck=false}
    \vspace{-1.5em}
    \caption{
    The framework of \textit{GigaWorld-0-Video-Dreamer}.}
    \label{fig:dreamer}
\end{figure}

\subsubsection{GigaWorld-0-Video-Dreamer}

\begin{figure}[!t]
    \centering
    \includegraphics[width=1\linewidth]{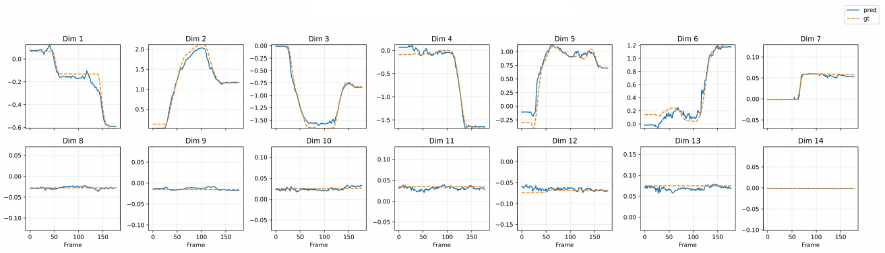}
    \captionsetup{type=figure, justification=justified, singlelinecheck=false}
\vspace{-1.5em}
    \caption{
    Qualitative comparison of action inference on the test set. \textcolor{blue}{Predicted} joint trajectories from \textit{GigaWorld-0-IDM} closely align with \textcolor{orange}{ground-truth} actions across all 12 arm joints and 2 gripper degrees of freedom, demonstrating high fidelity in recovering physically plausible manipulation policies from visual input alone.
    }
    \label{fig:idm}
\end{figure}

\textbf{Model Details.} \textit{GigaWorld-0-Video-Dreamer} is our foundation video generation model, capable of achieving IT2V generation. Its overall architecture is illustrated in Fig.~\ref{fig:dreamer}. We adopt a flow-matching~\citep{lipman2022flow} formulation for modeling the generative process:
\begin{equation}
    \frac{d\mathbf{z}_t}{dt} = \mathbf{v}_\theta(\mathbf{z}_t, t, \mathbf{c}),
\end{equation}
where $\mathbf{z}_t$ denotes the latent at time $t$, $\mathbf{c}$ represents the text and image conditioning, and $\mathbf{v}_\theta$ is the velocity parameterized by our model. For the input representation, we employ the 3D-VAE architecture~\citep{wan} to efficiently compress raw videos into latent representations with a spatial-temporal compression ratio of $4,8,8$ (temporal,height,width), resulting in 16-channel video latents. On top of this representation, we apply the same $1 \times 2 \times 2$ patchification strategy to further compress the latent features. We encode these latents using 3D Rotary Position Embedding (3D-RoPE)~\citep{rope}. For text conditioning, we utilize the T5 encoder~\citep{t5} to extract textual embeddings. The core generative backbone of \textit{GigaWorld-0-Video-Dreamer} is a DiT built upon sparse attention mechanisms~\citep{natten}. Additionally, we incorporate a Mixture-of-Experts (MoE) architecture~\citep{deepseekv2} into the feed-forward network (FFN) blocks of the DiT.  Let $u_t$ denote the FFN input of the $t$-th token, we compute the FFN output $\mathbf{h}_{t}^{\prime}$ as follows:
\begin{align}
\mathbf{h}_{t}^{\prime} &= \mathbf{u}_{t} +  \sum_{i=1}^{N_{r}} g_{i, t} \, \mathrm{FFN}_{i}\left(\mathbf{u}_{t}\right), \\
g_{i, t}^{\prime} &= 
\begin{cases}
s_{i, t}, & \text{if } s_{i, t} \in \operatorname{Topk}\left(\{s_{j, t} \mid 1 \leqslant j \leqslant N_{r}\}, K_{r}\right), \\
0, & \text{otherwise},
\end{cases} \\
s_{i, t} &= \operatorname{softmax}\left(\mathbf{u}_{t}^{\top} \mathbf{e}_{i}\right).
\end{align}
Specifically, in contrast to DeepSeek-V2~\citep{deepseekv2}, we do not include a shared expert. Instead, we configure $N_{r}=4$ routed experts and activate $K_{r}=2$ experts per token. This design enables dynamic specialization across different semantic regions of the video without redundant parameter sharing. $\mathbf{e}_{i}$ is the learnable vector of the $i$-th routed expert. Besides, to ensure MoE load balance, we employ a complementary balance loss from DeepSeek-V3~\citep{deepseekv3}:

\begin{align}
\mathcal{L}_{\text{Load}}&=\alpha \sum_{i=1}^{N_{r}} f_{i} P_{i}, \\
f_{i}&=\frac{N_{r}}{K_{r} T} \sum_{t=1}^{T} \mathds{1}\left(s_{i, t} \in \operatorname{Topk}\left(\left\{s_{j, t} \mid 1 \leqslant j \leqslant N_{r}\right\}, K_{r}\right)\right), \\
s_{i, t}^{\prime}&=\frac{s_{i, t}}{\sum_{j=1}^{N_{r}} s_{j, t}}, \\
P_{i}&=\frac{1}{T} \sum_{t=1}^{T} s_{i, t}^{\prime},
\end{align}
where $\alpha=0.01$ is a balance factor. $\mathds{1}(\cdot)$ denotes the indicator function, $T$ denotes the number of tokens in a video sequence. The balance loss encourages the expert load on each sequence to be balanced.


\textbf{Function as Data Engine.} \textit{GigaWorld-0-Video-Dreamer} serves as a versatile foundation model for training downstream controllable video generation systems. Moreover, it functions as a powerful data engine for training VLA models. 
As shown in Fig.~\ref{fig:videogen} and Fig.~\ref{fig:mv-videogen}, \textit{GigaWorld-0-Video-Dreamer} can generate distinct future videos conditioned on the same initial frame but guided by different text prompts. We then train an Inverse Dynamics Model, \textit{GigaWorld-0-IDM}, to infer the corresponding robotic arm actions from these generated videos.
Specifically, given a generated video sequence $\mathbf{V} = \{\mathbf{v}_1, \mathbf{v}_2, \dots, \mathbf{v}_T\}$, where $\mathbf{v}_t \in \mathbb{R}^{H \times W \times 3}$ denotes the RGB frame at time $t$, \textit{GigaWorld-0-IDM} estimates the joint-angle trajectory:
\begin{equation}
    \boldsymbol{\theta}_{1:T} = f_{\text{IDM}}(\mathbf{V}),
\end{equation}
where $\boldsymbol{\theta}_t = [\theta_t^{(1)}, \theta_t^{(2)}, \dots, \theta_t^{(D)}]^\top \in \mathbb{R}^D$ represents the rotation angles of all $D$ joints of the robotic arm at timestep $t$. 
In contrast to prior IDM~\citep{anypos}, the \textit{GigaWorld-0-IDM} employs {masked training. Specifically, we use~\citep{sam2} to segment the robotic arm from the input video and feed only the segmented arm region into the IDM during training, thereby reducing the adverse impact of cluttered backgrounds on prediction accuracy.  This strategy significantly enhances the model's robustness and prediction accuracy under real-world visual ambiguities. As shown in Fig.~\ref{fig:idm}, we collect manipulation data from unseen tasks for evaluation. \textit{GigaWorld-0-IDM} successfully infers action sequences that closely align with the ground-truth trajectories, accurately predicting the states of all 12 arm joints and the 2 gripper degrees of freedom.
The resulting paired dataset of generated videos and predicted actions $(\mathbf{V}, \boldsymbol{\theta}_{1:T})$ provides abundant, diverse, and temporally aligned supervision for training VLA models without requiring real-world robot interaction.

\begin{figure}[!t]
    \centering
    \includegraphics[width=0.8\linewidth]{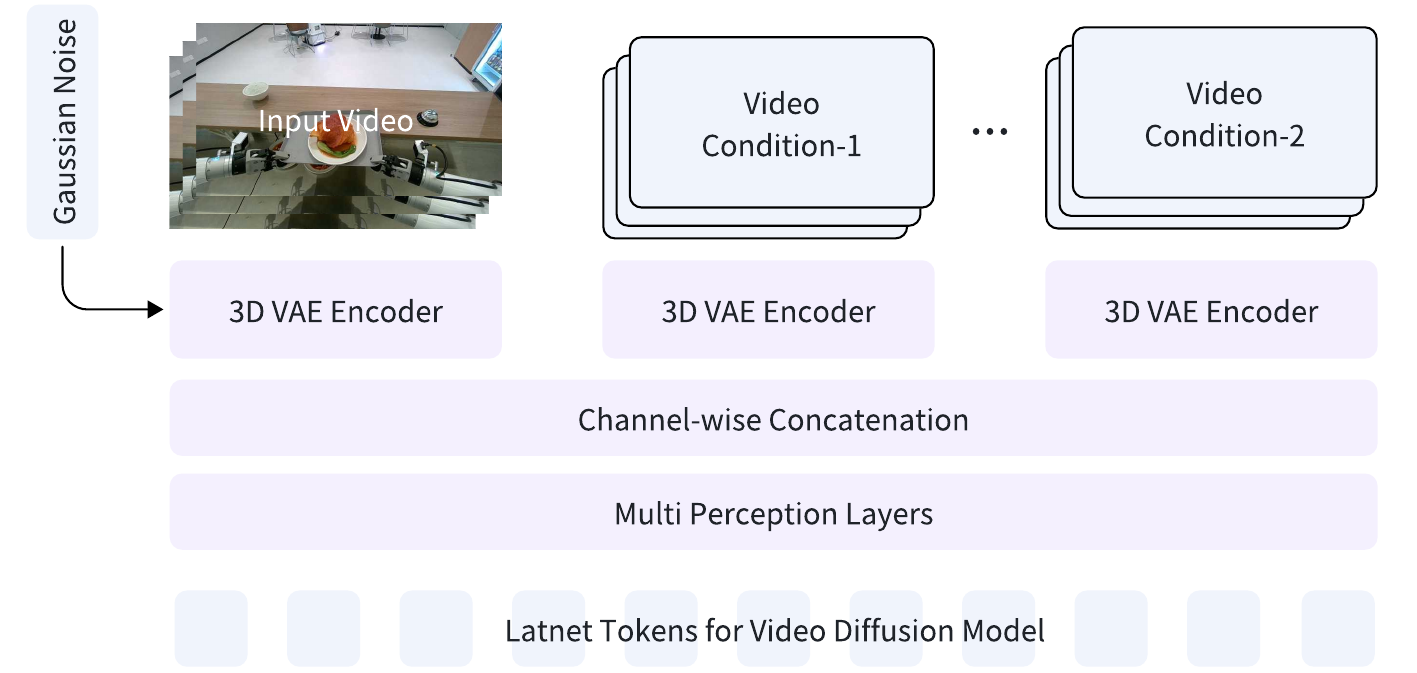}
    \captionsetup{type=figure, justification=justified, singlelinecheck=false}
    \caption{
    The control branch of \textit{GigaWorld-Video}.}
    \label{fig:control_branch}
\end{figure}

\subsubsection{GigaWorld-0-Video-AppearanceTransfer}

Real-world data collected for training VLA models often suffers from limited diversity in texture, color, and lighting conditions. This limitation hinders the robustness of trained models when deployed in complex, visually rich real-world environments. While traditional simulation pipelines can generate vast amounts of data with diverse textures, colors, and lighting, the rendered appearances still exhibit a significant sim2real gap, leading to low success rates in real robot deployment. To address this challenge, we propose \textit{GigaWorld-0-Video-AppearanceTransfer}, an efficient framework that enables text-driven appearance modification of real-world videos and facilitates style transfer from simulation to reality, thereby narrowing the sim2real gap.

\textbf{Model Details}. \textit{GigaWorld-0-Video-AppearanceTransfer} allows controllable editing of texture, material, and illumination in real-world video sequences using natural language prompts, while preserving geometric and motion consistency.
 Specifically, \textit{GigaWorld-0-Video-AppearanceTransfer} is built upon the pre-trained \textit{GigaWorld-0-Video-Dreamer}, extended with a lightweight control branch. As illustrated in Fig.~\ref{fig:control_branch}, ControlNet~\citep{controlnet} is not used due to its high parameter overhead, especially problematic when the base model employs a MoE architecture, where duplicating MoE layers would drastically increase model size. Instead, we introduce a more parameter-efficient control mechanism.
Given multiple video conditions (e.g., depth, surface normals), we first extract their latent representations using a 3D VAE~\citep{wan}. These control latents are then concatenated channel-wise with the noise latents used in the diffusion process. The combined tensor is passed through a series of channel-compressed MLP layers to produce the final latent input for the subsequent Transformer blocks. This design significantly reduces parameter count while remaining flexible across diverse video conditions.
For appearance control, we leverage textual prompts to independently manipulate foreground and background attributes such as texture, material, and lighting. To obtain geometric priors, we use VideoDepthAnything~\citep{videodepth} and LOTUS~\citep{lotus} to extract depth and normal maps from either real-world or simulation videos. These maps are normalized, repeated to form 3-channel inputs, and then encoded by the 3D VAE, following practices established in~\citep{emma,robotransfer}.

\textbf{Function as Data Engine}. \textit{GigaWorld-0-Video-AppearanceTransfer} functions as a powerful data engine for diverse visual generalization. As shown in Fig.~\ref{fig:apperance-transfer} and Fig.~\ref{fig:mv}, it supports both real2real and sim2real appearance transfer: given a real-world embodied interaction video or a simulation trajectory, the model can synthesize photorealistic variants with user-specified textures, colors, and lighting conditions via text prompts. This capability enables large-scale generation of visually diverse training data without requiring additional real-world collection or expensive simulation rendering.
VLA models trained on generated datasets demonstrate significantly improved robustness under appearance variations.

\subsubsection{GigaWorld-0-Video-ViewTransfer}

\begin{figure}[!t]
    \centering
    \includegraphics[width=1\linewidth]{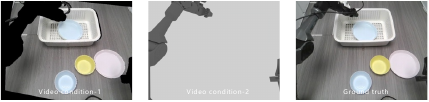}
    \captionsetup{type=figure, justification=justified, singlelinecheck=false}
    \caption{
    Training data pair of \textit{GigaWorld-0-Video-ViewTransfer}.}
    \label{fig:viewtransfer-datapair}
\end{figure}

Beyond appearance generalization, viewpoint generalization remains a critical challenge for VLA models. A model trained on data collected from viewpoint-A often fails to generalize to viewpoint-B. While multi-view data collection can mitigate this issue, it incurs prohibitive real-world annotation and operational costs. To address this, we propose \textit{GigaWorld-0-Video-ViewTransfer}, a framework that synthesizes diverse novel viewpoints from existing single-view robot interaction videos, while simultaneously transforming the associated robot actions to maintain task consistency.

\textbf{Model Details.} Formally, consider a robot operating in world coordinate frame $\mathcal{W}_A$, capturing an egocentric video $\mathbf{V}_A$ along with a sequence of end-effector poses $\{\mathbf{T}_t^{\text{ee} \rightarrow \text{base}}\}_{t=1}^T$ expressed relative to the robot base. We aim to synthesize a new observation $\mathbf{V}_B$ as if captured from a different world frame $\mathcal{W}_B$, where the robot base has been relocated (i.e., $\mathbf{T}^{\text{base}}_{\mathcal{W}_A \rightarrow \mathcal{W}_B} \neq \mathbf{I}$). Crucially, the absolute end-effector pose in the world frame must remain unchanged to preserve task semantics:
\begin{equation}
    \mathbf{T}_t^{\text{ee} \rightarrow \mathcal{W}} = \mathbf{T}^{\text{base} \rightarrow \mathcal{W}_A} \cdot \mathbf{T}_t^{\text{ee} \rightarrow \text{base}} = \mathbf{T}^{\text{base} \rightarrow \mathcal{W}_B} \cdot \mathbf{K}_t,
\end{equation}
where $\mathbf{K}_t$ denotes the new end-effector pose relative to the relocated base in $\mathcal{W}_B$. Solving for $\mathbf{K}_t$ yields:
\begin{equation}
    \mathbf{K}_t = \left( \mathbf{T}^{\text{base} \rightarrow \mathcal{W}_B} \right)^{-1} \cdot \mathbf{T}^{\text{base} \rightarrow \mathcal{W}_A} \cdot \mathbf{T}_t^{\text{ee} \rightarrow \text{base}}.
\end{equation}
The synthesized video $\mathbf{V}_B$ must be consistent with both the new camera viewpoint and the transformed action sequence $\mathbf{K} = \{\mathbf{K}_t\}_{t=1}^T$.

\textit{GigaWorld-0-Video-ViewTransfer} is built upon the pretrained \textit{GigaWorld-0-Video-Dreamer} via post-training adaptation, featuring a dual-condition control branch (Fig.~\ref{fig:control_branch}). To ensure 3D consistency under viewpoint changes, we decompose the control signal into two components:  
(i) background 3D consistency, enforced via video condition-1;  
(ii) robot arm 3D consistency, enforced via video condition-2.
Since paired multi-view real-world videos $(\mathbf{V}_A, \mathbf{V}_B)$ are unavailable, we employ a double-reprojection strategy~\citep{egodemogen} to construct self-supervised training pairs. For video condition-1, we first estimate scaled depth in $\mathcal{W}_A$ using MoGe~\citep{moge}, then warp $\mathbf{V}_A$ into the target view $\mathcal{W}_B$, and finally reproject it back to the original view (see video condition-1 in~Fig.\ref{fig:viewtransfer-datapair}). The reprojected video serves as the input condition, while the original $\mathbf{V}_A$ acts as the ground truth. To isolate scene geometry from the moving robot, we mask out the robotic arm~\citep{sam2} during the warping process.
For video condition-2, we render the transformed action sequence $\mathbf{K}$ in a physics-aware simulator~\citep{sapien} to generate an arm-only video that reflects the correct pose and kinematics in $\mathcal{W}_B$ (see video condition-2 in~Fig.\ref{fig:viewtransfer-datapair}). This rendered sequence provides explicit 3D guidance for arm motion consistency.
The model is trained to generate the ground-truth video $\mathbf{V}_A$ conditioned on both video condition-1 (background geometry under novel view) and video condition-2 (arm pose under transformed kinematics). More implementation details follow the similar framework ~\citep{egodemogen}.

\textbf{Function as Data Engine.} 
After training, \textit{GigaWorld-0-Video-ViewTransfer} functions as a scalable viewpoint augmentation engine. As shown in Fig.~\ref{fig:viewtransfer}, given a single real-world interaction video, it can generate photorealistic observations from arbitrarily novel viewpoints, accompanied by geometrically consistent robot actions $\mathbf{K}$. This enables massive expansion of the effective dataset with diverse egocentric perspectives, without additional real-world data collection. VLA models trained on this augmented data exhibit significantly improved robustness to viewpoint shifts during deployment, closing a key gap between simulation and real-world generalization.

\subsubsection{GigaWorld-0-Video-MimicTransfer}

\begin{figure}[!t]
    \centering
    \includegraphics[width=1\linewidth]{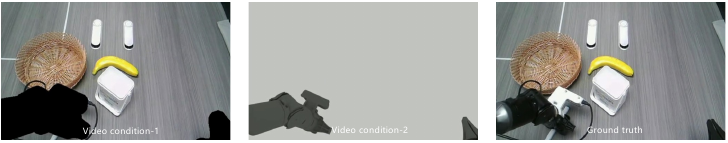}
    \captionsetup{type=figure, justification=justified, singlelinecheck=false}
    \caption{
    Training data pair of \textit{GigaWorld-0-Video-MimicTransfer}.}
    \label{fig:mimicdreamer_pair}
\end{figure}

Collecting real-world robot data via teleoperation is costly. An alternative and more efficient approach is to leverage first-person demonstration videos. However, a significant gap exists between first-person human demonstration videos and actual robot execution videos, the most prominent being the appearance gap between human hands and robotic arms. To address this issue, we propose \textit{GigaWorld-0-Video-MimicTransfer}, a method that translates first-person human-hand manipulation videos into robotic-arm manipulation videos, thereby reducing the aforementioned appearance gap and enhancing the usability of egocentric manipulation data.

\textbf{Model Details.} \textit{GigaWorld-0-Video-MimicTransfer} is a post-trained model based on \textit{GigaWorld-0-Video-Dreamer}, with its control branch illustrated in Fig.~\ref{fig:control_branch}. During training, due to the scarcity of aligned video pairs showing both human-hand and robotic-arm manipulations, we construct the training data using only robotic-arm manipulation videos. Specifically, we employ two video conditions: video condition-1 controls the manipulation scene, while video condition-2 enforces the robotic arm’s motion to mimic that of a human hand. As shown in Fig.~\ref{fig:mimicdreamer_pair}, to form video condition-1, we mask out the robotic arm in the original manipulation video and retain only the background. For video condition-2, we drive a simulated robotic arm using the original arm’s motion trajectories to generate a synthetic video depicting human-like manipulation. The model is trained to reconstruct the original (unmasked) robotic-arm manipulation video from these two conditions. Additional training details and data construction procedures follow those described in~\citep{mimicdreamer}.

\textbf{Function as a Data Engine.} After training, \textit{GigaWorld-0-Video-MimicTransfer} serves as a data engine that translates first-person human demonstration videos into robotic-arm manipulation videos, as shown in Fig.~\ref{fig:mimicdreamer}. Specifically, the human hand is masked out from the input video to serve as video condition-1, preserving the scene context. Meanwhile, using annotated end-effector poses of the human hand, we solve for the corresponding joint angles of the robotic arm via inverse kinematics (IK), and render the resulting arm pose in a simulator to generate video condition-2. Conditioned on these two inputs, the model synthesizes a realistic robotic-arm manipulation video that mimics the original human action, thereby enabling scalable and cost-effective data augmentation for robot learning.

\begin{figure}[t]
\centering
\captionsetup{type=figure, justification=justified, singlelinecheck=false}
\includegraphics[width=1\linewidth]{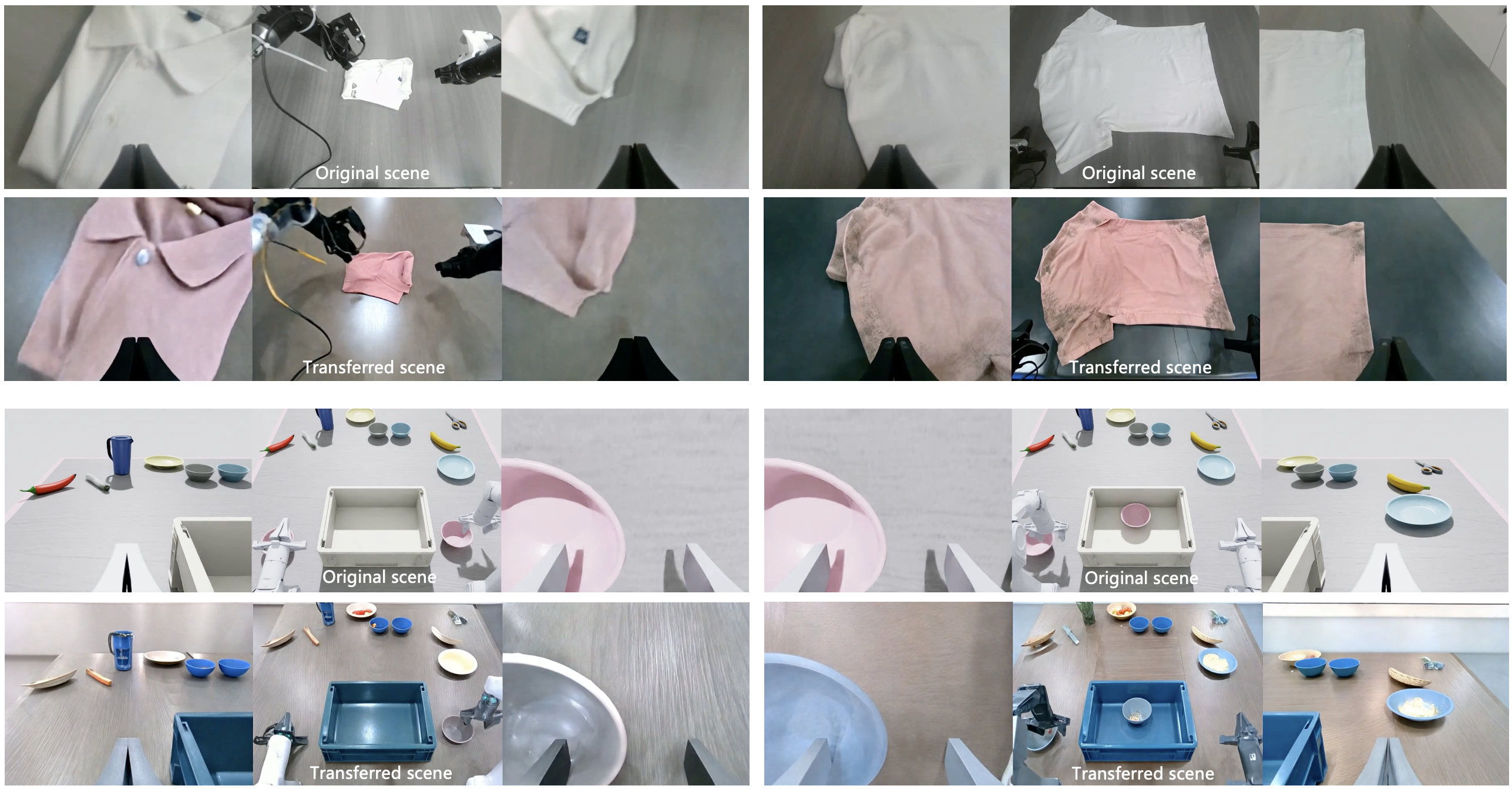}
\caption{GigaWorld can generate multi-view consistent videos, thereby enabling 3D-aware training and improving spatial reasoning in downstream tasks.}
\label{fig:mv}
\end{figure}

Furthermore, to better support embodied manipulation scenarios,\textit{GigaWorld-0-Video} incorporates multi-view video generation and generation acceleration techniques. For multi-view synthesis, we follow recent approaches~\citep{drivedreamer2,emma,robotransfer} by concatenating multi-view images along the width dimension into a single panoramic input. This design preserves the original diffusion model architecture and leverages in-context learning capabilities: after fine-tuning on a small set of multi-view data, the model generates temporally and spatially coherent videos across multiple viewpoints without architectural modifications. As shown in Fig.~\ref{fig:mv} and Fig.~\ref{fig:mv-videogen}, the resulting outputs exhibit strong cross-view consistency, making them suitable for training vision systems requiring egocentric and third-person observations.
To accelerate video generation, \textit{GigaWorld-0-Video} employs denoising step distillation~\citep{dmd2}, reducing the sampling process from dozens of steps to a single step. Combined with FP8-precision inference, these optimizations achieve over a $50\times$ speedup compared to standard diffusion models, enabling scalable data generation at deployment time.
Despite these advances, generated videos may still contain hallucinations or artifacts that could impair downstream policy learning. To ensure data quality, \textit{GigaWorld-0-Video} introduces a comprehensive evaluation pipeline that assesses each video across multiple dimensions: geometric consistency, multi-view coherence~\citep{robotransfer}, text-to-video alignment~\citep{cosmosreason}, and physical plausibility~\citep{cosmosreason}. A composite quality score is computed for each sequence, determining its suitability for pre-training, fine-tuning, or rejection.

\subsection{GigaWorld-0-3D}

While \textit{GigaWorld-0-Video} leverages video generation models to synthesize texture-rich embodied scene data, high-quality embodied manipulation also demands strong geometric consistency and physical accuracy. To address these requirements, \textit{GigaWorld-0-3D} adopts 3D Gaussian Splatting (3DGS)~\citep{3dgs} as its core scene representation, enabling the construction of spatially coherent and physically grounded 3D environments. Below, we detail the four key components of \textit{GigaWorld-0-3D}: \textit{GigaWorld-0-3D-FG}, \textit{GigaWorld-0-3D-BG}, \textit{GigaWorld-0-3D-Phys}, and \textit{GigaWorld-0-3D-Act}.

\subsubsection{GigaWorld-0-3D-FG}

\begin{figure}[h]
    \centering
    \includegraphics[width=1\linewidth]{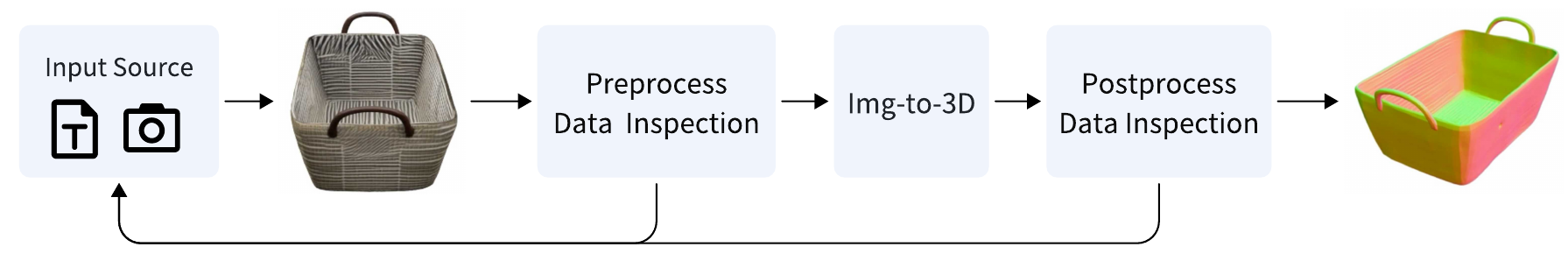}
    \captionsetup{type=figure, justification=justified, singlelinecheck=false}
    \caption{
    Overall pipeline of \textit{GigaWorld-0-3D-FG}.}
    \label{fig:GigaWorld-0-3d-fg}
\end{figure}

Traditional 3DGS reconstruction pipelines typically require dense multi-view inputs to produce high-fidelity assets. However, recent advances~\citep{clay,trellies,hunyuan3d1,hunyuan3d2,seed3d} in generative modeling have enabled object-level 3D reconstruction from extremely sparse inputs, such as a single image or even a text prompt. Notable approaches in this direction include Trellis and Hunyuan3D. Despite its strong geometric modeling capabilities, Trellis suffers from several shortcomings that limit its applicability in embodied AI scenarios: the generated textures often display poor visual fidelity, especially due to over-saturated specular highlights that lead to unnatural whitening when baked onto the mesh. Moreover, the resulting assets are purely visual constructs lacking real-world scale, physically plausible geometry, or material properties, rendering them incompatible with physics-based simulators~\citep{mujoco,issac,sapien}.

To bridge this gap, \textit{GigaWorld-0-3D-FG} enhances asset quality through rigorous data curation, while its successor module, \textit{GigaWorld-0-3D-Phys}, endows the assets with physical semantics suitable for simulation. As illustrated in Fig.~\ref{fig:GigaWorld-0-3d-fg}, the \textit{GigaWorld-0-3D-FG} pipeline accepts either a real-world photograph or a synthetic image generated via text-to-image models. Prior to 3D generation, an automated preprocessing stage performs quality control on the input. Specifically, we employ an aesthetic assessment module based on the Aesthetic-Checker~\citep{hpsv3}, which correlates positively with texture richness. Recognizing that foreground segmentation accuracy critically influences 3D output quality, we introduce an \textit{ImageSegChecker} powered by GPT-4o to evaluate segmentation reliability. To ensure robustness across diverse object categories, the system integrates three segmentation backends~\citep{sam2,remgb,remgb14}. If the \textit{ImageSegChecker} flags a segmentation failure, the pipeline triggers a retry, either by prompting the user to capture a new image or by regenerating the input using an alternative text-to-image model.

For the image-to-3D conversion, we adopt open-source generative models to facilitate seamless integration with future community advancements. Among available options, Trellis~\citep{trellies} is selected for its superior geometric coherence and its dual support for mesh and 3DGS representations. Following generation, a postprocessing inspection module, termed \textit{MeshGeoChecker}, renders the asset from four orthogonal viewpoints to evaluate geometric completeness and plausibility. Only assets that pass all quality gates are exported in URDF format and archived; those failing any inspection stage are automatically resubmitted to the corresponding generation step with modified parameters and random seeds for re-synthesis. Additional implementation details, including texture baking strategies, are provided in~\citep{embodiedgen}.

\subsubsection{GigaWorld-0-3D-BG}
\begin{figure}[h]
    \centering
    \includegraphics[width=1\linewidth]{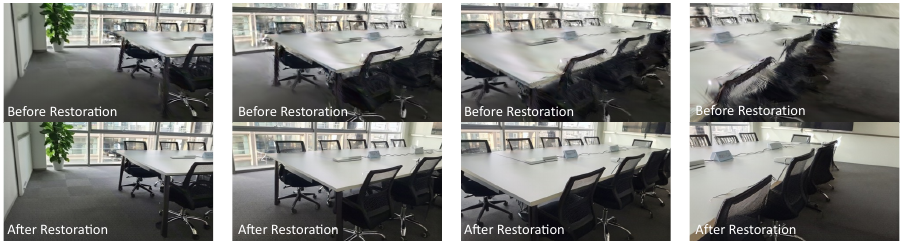}
    \captionsetup{type=figure, justification=justified, singlelinecheck=false}
    \caption{
    Visualization of novel view synthesis before and after view restoration.}
    \label{fig:bg}
\end{figure}

3DGS~\citep{3dgs} has emerged as a mature technique for scene reconstruction. Conventional 3DGS relies on Elliptical Weighted Average (EWA) splatting, which approximates projection via Jacobian computation and is limited to pinhole camera models. In contrast, 3DGRUT~\citep{3dgut} enhances camera compatibility by associating each 3D Gaussian with seven representative points (one center and six boundary points), enabling precise modeling of non-pinhole cameras, such as those with rolling shutters, commonly used in embodied AI settings. This leads to higher reconstruction fidelity in such scenarios.

However, traditional 3DGS typically requires dense multi-view inputs to achieve high-quality reconstructions, which are often unavailable in real-world embodied settings. To address this limitation, we draw inspiration from recent generative approaches~\citep{drivedreamer4d,recondreamer,recondreamer++,difix3d+} that synthesize novel views to enrich sparse observations. \textit{GigaWorld-0-3D-BG} pipeline begins with sparse-view inputs and employs 3DGRUT for initial scene reconstruction. Yet, under sparse-view conditions, novel view synthesis (NVS) often suffers from geometric and photometric artifacts.

To mitigate this, we adopt a view restoration strategy inspired by~\citep{recondreamer}, training a dedicated view refinement model to hallucinate plausible intermediate views. As shown in Fig.~\ref{fig:bg}, the refined views significantly reduce artifacts and provide dense, consistent visual observations. These synthesized views then serve as augmented inputs for a second-stage, dense 3DGS reconstruction, yielding a high-fidelity Gaussian representation of the background. Finally, we convert the resulting dense Gaussian Splats into a watertight mesh using the Poisson Surface Reconstruction, producing realistic and geometrically consistent background assets suitable for embodied manipulation scenarios.

\subsubsection{GigaWorld-0-3D-Phys}

\begin{figure}[h]
    \centering
    \includegraphics[width=1\linewidth]{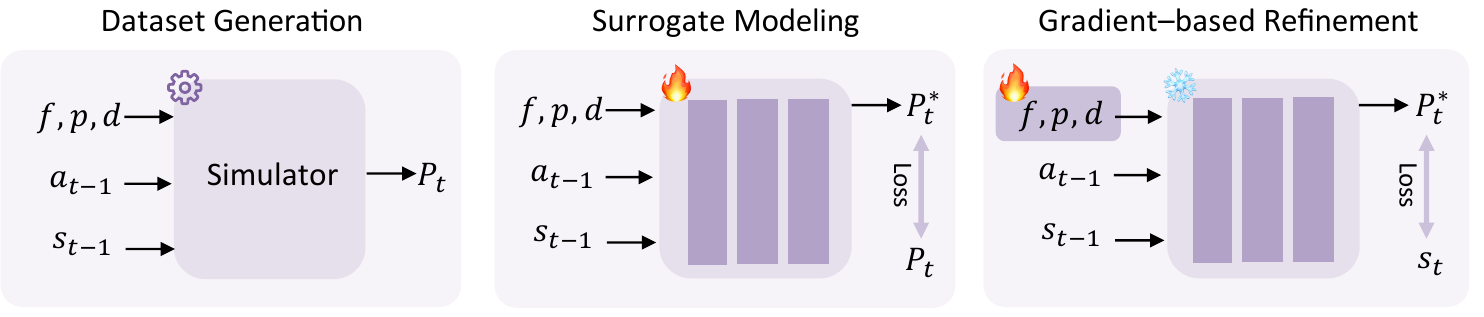}
    \captionsetup{type=figure, justification=justified, singlelinecheck=false}
    \caption{
    The learning pipeline of the differentiable physics network in \textit{GigaWorld-0-3D-Phys}.}
    \label{fig:phys}
\end{figure}

While \textit{GigaWorld-0-3D-FG} and \textit{GigaWorld-0-3D-BG} construct foreground and background assets using 3DGS and mesh representations, these assets lack physical properties necessary for interactive simulation. To enable physically grounded interaction, \textit{GigaWorld-0-3D-Phys} endows both robotic agents and manipulated objects with realistic physical attributes.

For the robotic arm, precise physical parameters, such as joint friction and PD controller gains, are critical but rarely measurable in practice. Traditional system identification methods (e.g., manual tuning or simulated annealing~\citep{simplerenv}) are slow and labor-intensive. In contrast, \textit{GigaWorld-0-3D-Phys} leverages a differentiable physics framework based on physics-informed neural networks (PINNs), enabling efficient, gradient-based parameter estimation. The pipeline operates in three stages, as shown in Fig.~\ref{fig:phys}:  
(1) Real-world trajectories $(\mathbf{a}_{t-1}, \mathbf{s}_{t-1})$ are paired with randomly sampled physical parameters $(f, p, d)$, denoting friction, stiffness, and damping, and used to generate simulated rollouts.  
(2) A surrogate model $\mathcal{M}_{f,p,d}$ is trained to approximate the simulator’s dynamics by minimizing the MSE between predicted and simulated next states, yielding a differentiable dynamics model.  
(3) With the surrogate model fixed, physical parameters are refined via gradient descent to minimize the discrepancy between simulated and real trajectories, converging to an optimal set $(f^*, p^*, d^*)$ that accurately replicates real-world behavior. Further details are provided in~\citep{embodiedreamer}.

For manipulated objects, we employ a multimodal physics expert agent built on Qwen3-VL~\citep{qwen3} that infers physical properties from rendered orthographic views. The agent first estimates real-world scale by analyzing a frontal view under text-guided constraints, resolving ambiguities in object size. Once scaled, it predicts mass, friction coefficient, and other physical attributes, associating them with semantic categories for downstream use, refer to~\citep{embodiedgen} for more implementation details.

For deformable objects, we extend the 3DGS representation by binding spring-mass systems to Gaussian particles, following the spirit of PhysTwin~\citep{phystwin}. However, unlike PhysTwin’s per-scenario optimization, we are exploring a feedforward approach that directly infers spring-mass parameters from monocular video, enabling fast, generalizable soft-body simulation.

\subsubsection{GigaWorld-0-3D-Act}

\begin{figure}[!t]
    \centering
    \includegraphics[width=0.9\linewidth]{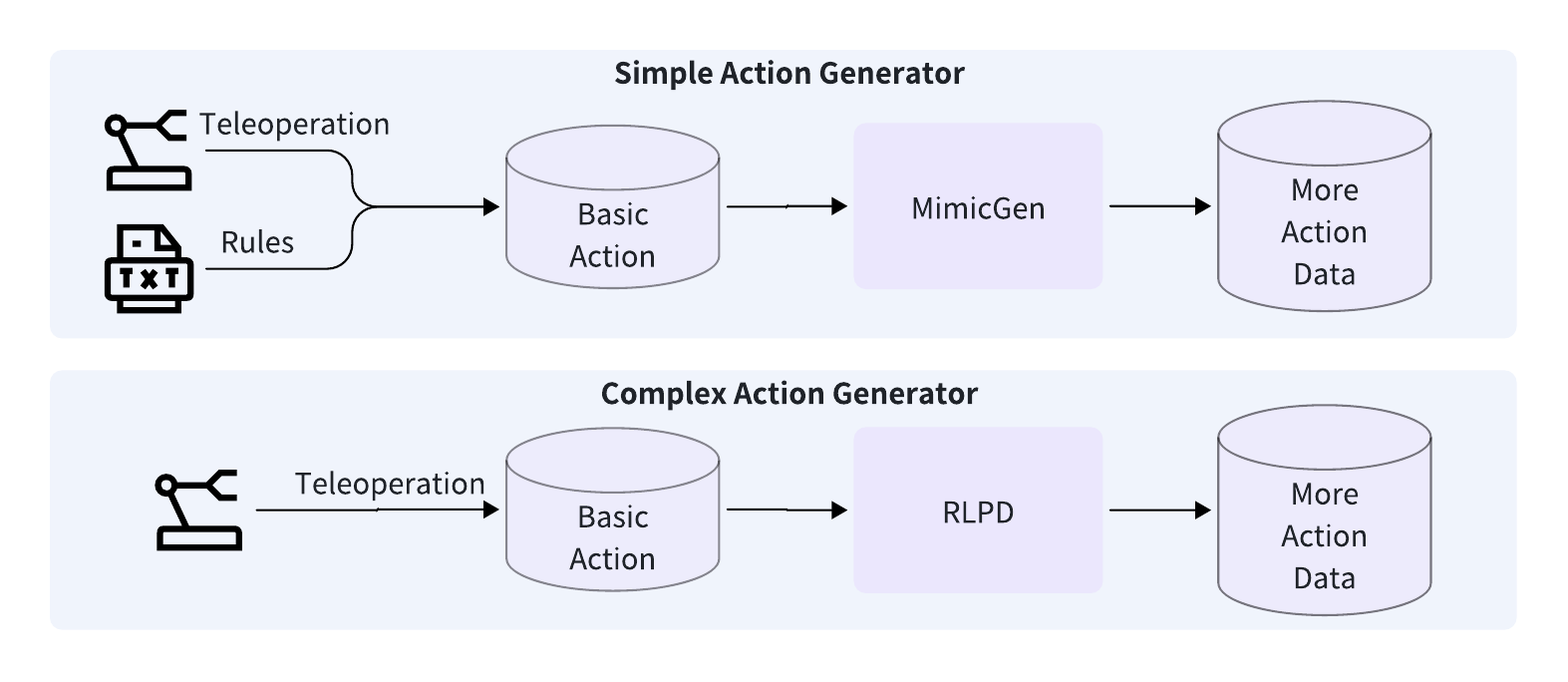}
    \captionsetup{type=figure, justification=justified, singlelinecheck=false}
    \caption{
    The overall pipeline of \textit{GigaWorld-0-3D-Act}.}
    \label{fig:act}
\end{figure}

While the community has developed numerous simulation-based approaches for automatic action generation, enabling task execution and large-scale expansion of robotic manipulation datasets, these methods often lack adaptability across scene complexity or generalization to novel object configurations. To address this, \textit{GigaWorld-0-3D-Act} introduces a two-tiered action generation pipeline tailored to both simple and complex manipulation scenarios, as shown in Fig.~\ref{fig:act}.

For \textbf{simple scenarios}, \textit{GigaWorld-0-3D-Act} first acquires a small set of basic demonstrations via teleoperation or rule-based policies~\citep{robotwin}. These seed trajectories are then systematically extended to new object poses and scene layouts using the \textit{MimicGen} framework~\citep{mimicgen}, enabling scalable and geometrically consistent action augmentation without additional human supervision.

For \textbf{complex scenarios} involving multi-step reasoning or contact-rich interactions, \textit{GigaWorld-0-3D-Act} leverages teleoperated demonstrations as cold-start data for reinforcement learning. It then employs fast online reinforcement learning (e.g., RLPD~\citep{rlpd}) to rapidly bootstrap policy training. Once converged, the learned policy is deployed to generate large-scale, physically plausible, and diverse manipulation trajectories, effectively bridging high-level task structure with low-level motor control.

\subsubsection{Function as Data Engine}
The \textit{GigaWorld-0-3D} model suite, when integrated as a cohesive pipeline, constructs geometrically consistent and physically realistic embodied manipulation scenes that serve as high-quality training data for VLA models, as shown in Fig.~\ref{fig:gigaworld3d}. Specifically, \textit{GigaWorld-0-3D-FG} and \textit{GigaWorld-0-3D-BG} generate foreground and background assets in complementary representations: 3DGS for photorealistic rendering, and meshes for accurate collision detection, dynamics simulation, and physical interaction. \textit{GigaWorld-0-3D-Phys} then endows both the robotic manipulator and scene objects with physically grounded properties (e.g., mass, friction, elasticity) and performs differentiable system identification to calibrate actuation dynamics. Finally, \textit{GigaWorld-0-3D-Act} synthesizes executable, collision-free manipulation trajectories that complete user-specified tasks. The resulting rendered data are directly usable for end-to-end VLA training.
To further enhance data diversity, we integrate \textit{GigaWorld-0-Video-AppearanceTransfer} to perform text-guided editing of texture, color, material, and lighting across the rendered scenes. This enables zero-shot domain expansion while preserving geometric and physical consistency, dramatically increasing the visual and contextual richness of the training distribution without additional 3D asset creation.

\section{GigaWorld-0 Training}

Our training data combines publicly available datasets with proprietary data collected from our in-house robotic platforms. Public sources include {AgiBotWorld}~\citep{agibot} and {RoboMind}~\citep{robomind}, which provide foundational coverage of manipulation and locomotion tasks. In addition, we collected thousands of hours of proprietary data using the Agilex Cobot Magic and AgiBot G1 platforms across a total area of 3,100~m$^2$, spanning five broad environment categories: industrial, commercial, office, residential, and laboratory settings. These are further subdivided into 14 distinct real-world scenarios, including supermarkets, hotel lobbies, coffee shops, bubble tea stores, convenience stores, restaurants, warehouse material handling zones, industrial assembly lines, pantries, private residences, apartment interiors, meeting rooms, office workstations, and laboratories. The collected tasks range from basic pick-and-place operations to long-horizon sequential activities, mobile manipulation in dynamically changing layouts, and interactions with deformable objects.

Training video foundation models is computationally intensive. To enable efficient and cost-effective training, \textit{GigaWorld-0-Video-Dreamer} adopts sparse attention and FP8-precision training. Moreover, since most contemporary VLA models~\citep{pi0,pi05,gigabrain} operate on 480p inputs, we train our model at a resolution of 480$\times$768 for 61-frame sequences, striking a balance between visual fidelity and training efficiency.

Our training infrastructure, \textit{GigaTrain}\footnote{\url{https://github.com/open-gigaai/giga-train}}~\citep{gigatrain}, is a unified distributed framework designed for scalability and flexibility. It supports seamless multi-GPU/multi-node execution and integrates leading large-model training strategies, including:
\begin{itemize}
    \item Distribution framework: DeepSpeed ZeRO (Stages 0–3), FSDP2;
    \item Mixed-precision training (FP16, BF16, FP8);
    \item Gradient accumulation, gradient checkpointing, and exponential moving average (EMA);
    \item Configurable optimizers, learning rate schedulers, and other training modules.
\end{itemize}
This design enables both large-scale pretraining and resource-constrained post-training (e.g., fine-tuning with limited compute). To facilitate community adoption, we report resource consumption for various post-training configurations under modest hardware (e.g., 8$\times$H20 GPUs with batch size 32), providing practical guidance for users seeking to adapt our model and framework for downstream embodied tasks.

\begin{table}[t]
\centering
\small
\captionsetup{type=table, justification=justified, singlelinecheck=false}
\caption{Training efficiency of \textit{GigaWorld-0-Video-Dreamer} under different distributed training configurations (8$\times$H20 GPUs, batch size 32). Act. Ckpt. indicates activation checkpoint of the feedforward network. OOM. = Out of Memory.}
\label{tab:training_configs}
\begin{tabular}{@{}l c c c c c c@{}}
\toprule
Dist. Framework&  Act. Ckpt. & FP8 & Sparse Attn. & w/ MoE & Time (s/step) & Memory (MB) \\
\midrule
DeepSpeed-Zero0 &   &   &   &   & OOM. & OOM. \\
DeepSpeed-Zero2 &   &   &   &   & 32.84 & 95241 \\
FSDP-2          &   &   &   &   & 33.19 & 89355 \\

\midrule
DeepSpeed-Zero0 &  &  \cmark &   &   & 29.61 & 89781 \\
DeepSpeed-Zero2 &  & \cmark  &   &   & 29.75 & 76419 \\
FSDP-2          &  &  \cmark &   &   & 29.53 & 71857 \\

\midrule
DeepSpeed-Zero0 &  & \cmark &  \cmark &   & 25.54 & 90077 \\
DeepSpeed-Zero2 &  & \cmark & \cmark  &   & 25.44 & 76937 \\
FSDP-2          &  & \cmark &  \cmark &   & 25.38 & 73131 \\

\midrule
DeepSpeed-Zero0 & \cmark & \cmark & \cmark & \cmark & OOM. & OOM. \\
DeepSpeed-Zero2 & \cmark & \cmark & \cmark & \cmark & 33.27 & 84699 \\
FSDP-2          & \cmark & \cmark & \cmark & \cmark & 33.38 & 73997 \\
\bottomrule
\end{tabular}
\end{table}

As shown in Tab.~\ref{tab:training_configs}, FSDP-2 achieves the best memory efficiency among distributed training frameworks, followed by DeepSpeed ZeRO-2 and then ZeRO-0, though stronger memory optimization comes at the cost of increased communication overhead and slightly longer per-step latency. The use of FP8 precision consistently reduces both memory consumption and training time across all frameworks, demonstrating its effectiveness for scalable video foundation model training. For attention efficiency, we adopt NATTEN~\citep{natten} as our sparse attention operator due to its superior speedup over SageAttention~\citep{sageattn}, though it requires fine-tuning to avoid performance degradation in the absence of adaptation. Finally, when scaling to a 4-expert MoE architecture, the increased parameter footprint necessitates the application of activation checkpointing, specifically on the feedforward networks, to maintain feasible memory usage during training, enabling stable convergence under constrained hardware.
\section{Experiments}

\begin{figure}[!t]
    \centering
    \includegraphics[width=1\linewidth]{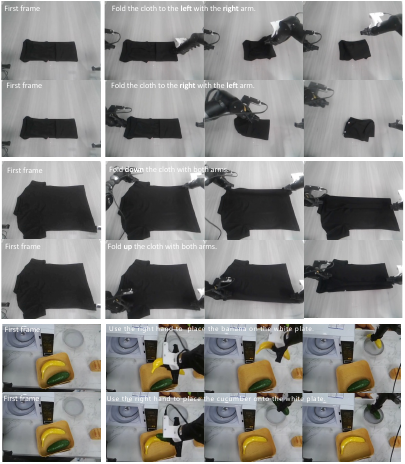}
    \captionsetup{type=figure, justification=justified, singlelinecheck=false}
    \caption{
    Visualization results of \textit{GigaWorld-0-Video-Dreamer} conditioned on the same initial frame but different text prompts, demonstrating its ability to produce diverse, semantically consistent future trajectories.}
    \label{fig:videogen}
\end{figure}

\begin{figure}[!t]
    \centering
    \includegraphics[width=1\linewidth]{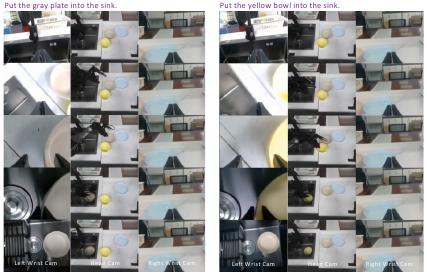}
    \captionsetup{type=figure, justification=justified, singlelinecheck=false}
    \caption{
    Multi-view visualization results of \textit{GigaWorld-0-Video-Dreamer} conditioned on the same initial frame but different text prompts.}
    \label{fig:mv-videogen}
\end{figure}

\begin{figure}[!t]
    \centering
    \includegraphics[width=1\linewidth]{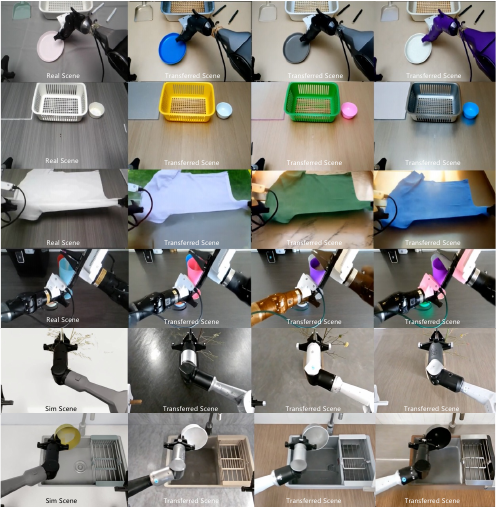}
    \captionsetup{type=figure, justification=justified, singlelinecheck=false}
    \caption{
    Visualization results of \textit{GigaWorld-0-Video-AppearanceTransfer}, which enables photorealistic editing of texture, material, and lighting in real-world or simulation-acquired videos while preserving scene geometry, object semantics, and temporal coherence.}
    \label{fig:apperance-transfer}
\end{figure}

To evaluate the capabilities of our foundation model \textit{GigaWorld-0-Video-Dreamer}, we conduct comprehensive assessments on two benchmarks (DreamGen Bench~\citep{dreamgen}, PBench~\citep{cosmosp2}), measuring performance across multiple dimensions: physical plausibility, geometric consistency, text-to-video alignment, multi-view coherence, and visual fidelity. Quantitative results demonstrate that \textit{GigaWorld-0-Video-Dreamer} achieves state-of-the-art performance in almost all metrics, significantly outperforming existing video generation models in simulating realistic embodied interactions.
In addition to quantitative evaluation, we perform extensive visual analysis to validate the semantic correctness, temporal coherence, and contextual richness of the generated videos. These visualizations confirm that \textit{GigaWorld-0} produces physically grounded and spatially consistent scenes under diverse object configurations and environmental conditions.
Ultimately, as a scalable data engine for embodied AI, \textit{GigaWorld-0} demonstrates strong downstream utility. Policies trained on \textit{GigaWorld-0}-generated data exhibit significant improvements in VLA task performance, particularly in generalization scenarios involving novel textures, novel object placements, and novel camera viewpoints.

\begin{table}[t]
\centering
\small
\setlength{\tabcolsep}{3pt} 
\captionsetup{type=table, justification=justified, singlelinecheck=false}
\caption{Evaluation on PBench Robot Set~\citep{cosmosp2}. For each column, the highest score is \textbf{bolded}.}
\label{tab:pbench}
\begin{tabular}{@{}l l c c c c c c c c c c@{}}
\toprule
\multirow{2}{*}{Model} & \multirow{2}{*}{\#Param.} 
& \multicolumn{8}{c}{Quality Score} & \multirow{2}{*}{Domain} & \multirow{2}{*}{Overall} \\
\cmidrule(lr){3-10} 
& 
& i2v-bg & i2v-s & aes & img & bg-con & mot & sub-con & o-con & Score & Score \\
\midrule
Cosmos-Predict2        & 14B & 97.4 & 97.3 & 47.0 & \textbf{94.0} & 66.3 & 98.9 & 12.0 & \textbf{93.1} & 84.0 & 79.88 \\
Wan2.2                 & 14B & 95.9 & 95.9 & 48.6 & 91.0 & 67.1 & 97.6 & 11.9 & 88.1 & 83.2 & 78.85 \\
Wan2.2                 & 5B  & 95.4 & 95.0 & 46.7 & 92.7 & 63.9 & 97.9 & 12.0 & 90.2 & 80.1 & 77.15 \\
Cosmos-Predict2.5      & 2B  & 93.8 & 91.3 & \textbf{49.3} & 92.1 & \textbf{74.2} & \textbf{99.2} & 12.2 & 90.2 & 84.7 & 79.95 \\
\textbf{GigaWorld-0-Video-Dreamer} & \textbf{2B(Act.)} & \textbf{97.6} & \textbf{97.6} & {48.1} & {93.6} & {66.8} & \textbf{99.2} & \textbf{12.6} & {91.9} & \textbf{88.2} & \textbf{82.07} \\
\bottomrule
\end{tabular}
\end{table}

\begin{table}[h]
\centering
\small
\setlength{\tabcolsep}{4pt}
\captionsetup{type=table, justification=justified, singlelinecheck=false}
\caption{Evaluation on DreamGen Bench~\citep{dreamgen}. For each column, the highest score is \textbf{bolded}.}
\label{tab:dreamgenbench}
\resizebox{1\textwidth}{!}{
\begin{tabular}{@{}l l 
  *{3}{c} 
  *{3}{c} 
  *{3}{c}@{}}
\toprule
\multirow{2}{*}{Method} & \multirow{2}{*}{Param.} 
& \multicolumn{3}{c}{GR1-Env} 
& \multicolumn{3}{c}{GR1-Object} 
& \multicolumn{3}{c}{GR1-Behavior} \\
\cmidrule(lr){3-5} \cmidrule(lr){6-8} \cmidrule(l){9-11}
& 
& Qwen-IF & GPT-IF & PA 
& Qwen-IF & GPT-IF & PA
& Qwen-IF & GPT-IF & PA\\
\midrule
Cosmos-Predict2        & 14B & \textbf{0.966} & {0.552} & \textbf{0.586} & {0.840} & {0.760} & 0.471 & \textbf{0.894} & \textbf{0.638} & 0.458 \\
Wan2.2                 & 14B & 0.900 & \textbf{0.760} & {0.549} & 0.700 & \textbf{0.780} & \textbf{0.531} & {0.870} & {0.570} & \textbf{0.477} \\
Wan2.2                 & 5B  & 0.790 & 0.340 & 0.531 & 0.720 & \textbf{0.780} & 0.522 & 0.830 & 0.280 & 0.468 \\
Cosmos-Predict2.5      & 2B  & {0.930} & 0.480 & 0.534 & \textbf{0.920} & 0.240 & {0.503} & 0.830 & 0.320 & {0.471} \\
\textbf{GigaWorld-0-Video-Dreamer} & \textbf{2B(Act.)} & \textbf{0.966} & {0.586} & 0.529 & \textbf{0.920} & 0.540 & 0.481 & \textbf{0.894} & \textbf{0.638} & 0.446 \\
\bottomrule
\end{tabular}}
\end{table}

\subsection{Benchmark Results}

While a wide variety of world model benchmarks~\citep{cosmosp2,dreamgen,vbench,worldinworld,worldscore} exist, spanning general-purpose scene understanding, content creation, embodied navigation, and embodied manipulation, we specifically select PBench~\citep{cosmosp2} and DreamGen Bench~\citep{dreamgen} for evaluation, as they are explicitly designed for embodied manipulation tasks and provide comprehensive metrics for assessing visual quality, physical plausibility, geometric consistency.

As shown in Tab.~\ref{tab:pbench}, we compare our model against state-of-the-art video generation approaches~\citep{wan,cosmosp2}, including Cosmos-Predict2-14B, Cosmos-Predict2.5-2B, and Wan2.2-5B and Wan2.2-14B. Despite having the smallest activated parameter, \textit{GigaWorld-0-Video-Dreamer} achieves the highest overall score on PBench (Robot Set), demonstrating superior efficiency and generation quality for embodied AI applications.

Additionally, we provide detailed evaluation on {DreamGenBench}. Following the official protocol, we fine-tune both our model and open-source baselines (Wan2.2 and Cosmos-Predict2/2.5) on the publicly released GR1 robot dataset, which comprises 29 sequences for GR1-Env, 50 for GR1-Obj, and 47 for GR1-Behavior. All models are fine-tuned using the same hyperparameters as in DreamGen~\citep{dreamgen}: batch size 64 for 200 training steps.
During evaluation, we use the official DreamGen Bench codebase. Notably, the prompts embedded in the publicly released code differ from those reported in the original paper, to ensure fair comparison, we adopt the exact prompts described in the DreamGen paper~\citep{dreamgen}, which enables us to reproduce their reported results. Furthermore, following the benchmark’s instructions, we compute the PA II score using the VideoPhy~\citep{videophy} protocol and report the final PA score as the average of PA I~\citep{dreamgen} and PA II.
Results on DreamGen Bench are shown in Tab.~\ref{tab:dreamgenbench}. Despite the absence of extensive GR1 data in our pretraining corpus, \textit{GigaWorld-0-Video-Dreamer}, with only 2B activated parameters, consistently outperforms the comparable Cosmos-Predict2.5-2B across all three scenarios (GR1-Env, GR1-Obj, and GR1-Behavior) in terms of instruction-following fidelity, demonstrating stronger generalization and controllability in embodied manipulation.

\begin{figure}[!t]
    \centering
    \includegraphics[width=1\linewidth]{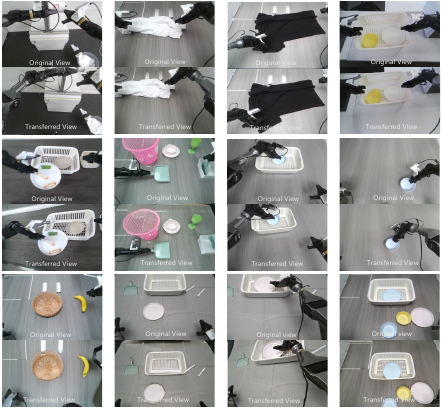}
    \captionsetup{type=figure, justification=justified, singlelinecheck=false}
    \caption{
    Visualization results of \textit{GigaWorld-0-Video-ViewTransfer}, which synthesizes photorealistic videos from arbitrary camera viewpoints while simultaneously adapting robot arm trajectories to maintain physical plausibility and action consistency, enabling the generation of diverse embodied manipulation data.}
    \label{fig:viewtransfer}
\end{figure}

\begin{figure}[!t]
    \centering
    \includegraphics[width=1\linewidth]{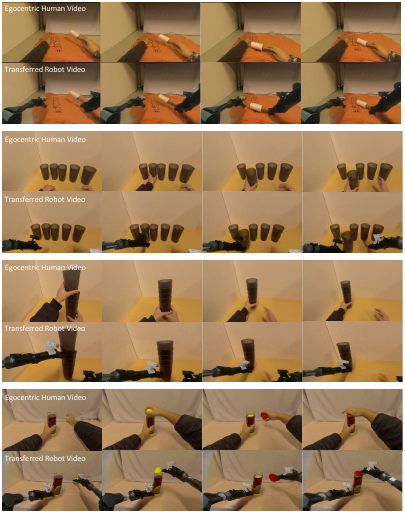}
    \captionsetup{type=figure, justification=justified, singlelinecheck=false}
    \caption{
    Visualization results of \textit{GigaWorld-0-Video-MimicTransfer}, which translates first-person human demonstration videos into robot-executable manipulation trajectories, enabling scalable synthesis of cross-embodiment training data for VLA models.}
    \label{fig:mimicdreamer}
\end{figure}

\begin{figure}[!t]
    \centering
    \includegraphics[width=1\linewidth]{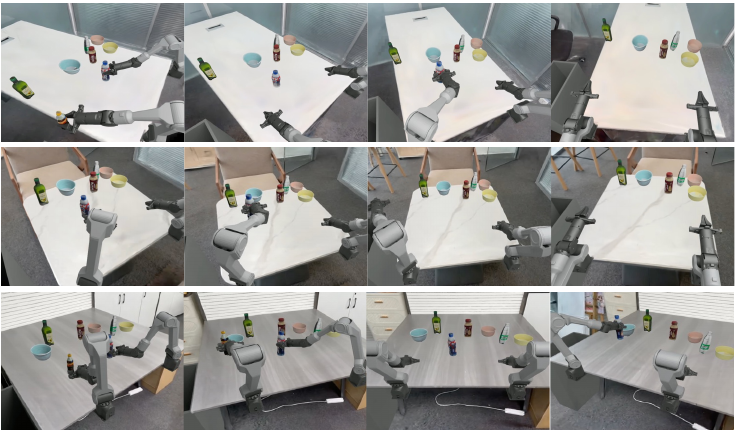}
    \captionsetup{type=figure, justification=justified, singlelinecheck=false}
    \caption{
    Visualization results of \textit{GigaWorld-0-3D}, showcasing geometrically consistent rendering and physically realistic robot actions.}
    \label{fig:gigaworld3d}
\end{figure}

\subsection{Visualiztaion Results}
Visualization serves as a crucial qualitative tool for evaluating world models, offering direct insight into the fidelity, controllability, and physical plausibility of generated data. To this end, we conduct extensive visual analysis of videos synthesized by \textit{GigaWorld-0}. 
As shown in Fig.~\ref{fig:videogen}, \textit{GigaWorld-0-Video-Dreamer} generates diverse future trajectories from a shared initial frame under varying text prompts, ranging from rigid-body manipulation (e.g., grasping objects) to deformable object interactions (e.g., folding laundry). The results demonstrate strong adherence to textual instructions, high visual fidelity, and stable temporal coherence across complex scenes. Additional video examples are available on our project page.
Moreover, \textit{GigaWorld-0-Video-Dreamer} supports multi-view video generation, as illustrated in Fig.~\ref{fig:mv-videogen}. The rendered views exhibit consistent geometry, appearance, and action dynamics across camera poses, while faithfully following high-level instructions. Such multi-view consistency is essential for training VLA policies, which typically rely on multi-perspective observations. Empirically, we find that IDMs used for gripper state estimation critically depend on multi-view inputs to achieve accurate and robust predictions, highlighting the practical value of our multi-view generation capability.

We further visualize the capabilities of \textit{GigaWorld-0-Video-AppearanceTransfer}, as shown in Fig.~\ref{fig:apperance-transfer}. This module enables photorealistic, text-guided editing of texture, material, and lighting in both real-world collected videos and simulation-generated sequences. By transferring rich, diverse appearance attributes while preserving motion dynamics and geometric structure, it facilitates efficient, low-cost augmentation of training data for embodied AI, significantly expanding the visual and contextual diversity of the training distribution without requiring new physical recordings or 3D asset authoring.

Additionally, Fig.~\ref{fig:viewtransfer} demonstrates the capability of \textit{GigaWorld-0-Video-ViewTransfer} to synthesize photorealistic observations from arbitrary camera viewpoints, even when conditioned on real-world collected videos. This enables seamless augmentation of single-view datasets with multi-perspective analogs, significantly enriching the spatial diversity of training data without requiring additional physical captures or costly 3D reconstruction.

Furthermore, Fig.~\ref{fig:mimicdreamer} illustrates the effectiveness of \textit{GigaWorld-0-Video-MimicTransfer} in translating first-person human manipulation demonstrations into robot-executable trajectories. The generated motions exhibit precise spatial alignment between the human hand and the robot end-effector, while preserving natural dynamics and physical plausibility. This cross-embodiment transfer greatly enhances the utility of egocentric video data for training VLA models, enabling the use of abundant, low-cost human demonstration videos as a scalable supervision signal for robotic policy learning.

Finally, we visualize the 3D scenes generated by \textit{GigaWorld-0-3D}, as shown in Fig.~\ref{fig:gigaworld3d}. The foreground objects are synthesized by \textit{GigaWorld-0-3D-FG}, while the background is reconstructed from real-world captures using \textit{GigaWorld-0-3D-BG} via 3DGS and mesh-based modeling. The resulting scene exhibits high geometric consistency across viewpoints and seamless integration between foreground and background components. Combined with physically plausible object properties and articulated dynamics from \textit{GigaWorld-0-3D-Phys} and \textit{GigaWorld-0-3D-Act}, this enables the construction of photorealistic, spatially coherent, and simulation-ready embodied manipulation environments—providing a robust foundation for training and evaluating VLA models.

\begin{figure}[htbp]
\centering
\captionsetup{type=figure, justification=justified, singlelinecheck=false}
\includegraphics[width=1\linewidth]{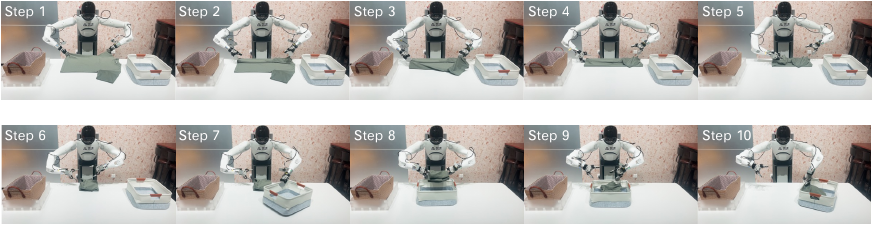}
\caption{Deployment of \textit{GigaBrain-0} on the G1 humanoid robot for real-world \texttt{laundry folding}.}
\label{fig:laundry_demo}
\end{figure}

\begin{figure}[htbp]
\centering
\captionsetup{type=figure, justification=justified, singlelinecheck=false}
\includegraphics[width=1\linewidth]{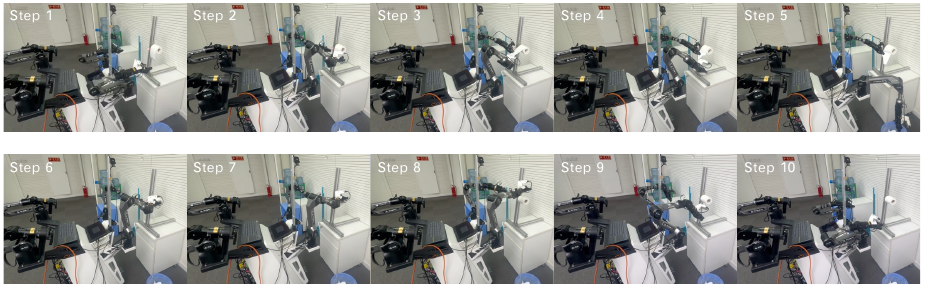}
\caption{Deployment of \textit{GigaBrain-0} on the PiPER arms for real-world \texttt{paper towel preparation}.}
\label{fig:papertowel_demo}
\end{figure}

\begin{figure}[htbp]
\centering
\captionsetup{type=figure, justification=justified, singlelinecheck=false}
\includegraphics[width=1\linewidth]{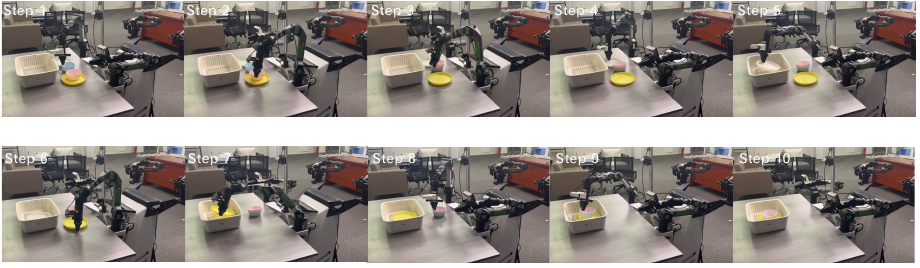}
\caption{Deployment of \textit{GigaBrain-0} on PiPER arms for real-world \texttt{table bussing}.}
\label{fig:table_demo}
\end{figure}

\begin{figure}[htbp]
\centering
\captionsetup{type=figure, justification=justified, singlelinecheck=false}
\includegraphics[width=1\linewidth]{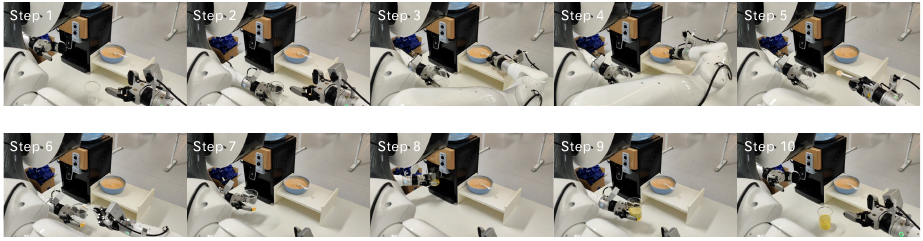}
\caption{Deployment of \textit{GigaBrain-0} on G1 humanoid robot for real-world \texttt{juice preparation}.}
\label{fig:juice_demo}
\end{figure}

\begin{figure}[htbp]
\centering
\captionsetup{type=figure, justification=justified, singlelinecheck=false}
\includegraphics[width=1\linewidth]{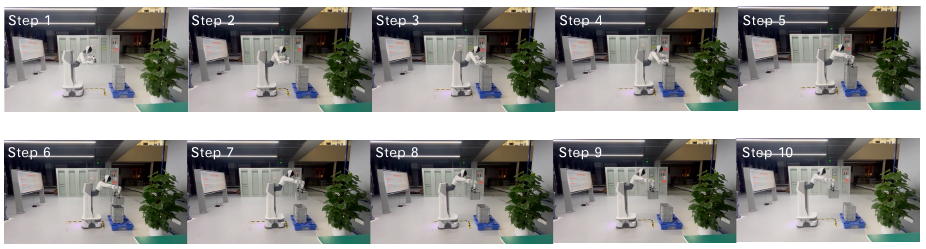}
\caption{Deployment of \textit{GigaBrain-0} on the G1 humanoid robot for real-world \texttt{paper towel preparation}.}
\label{fig:boxes_demo}
\end{figure}

\begin{figure}[htbp]
\centering
\captionsetup{type=figure, justification=justified, singlelinecheck=false}
\includegraphics[width=1\linewidth]{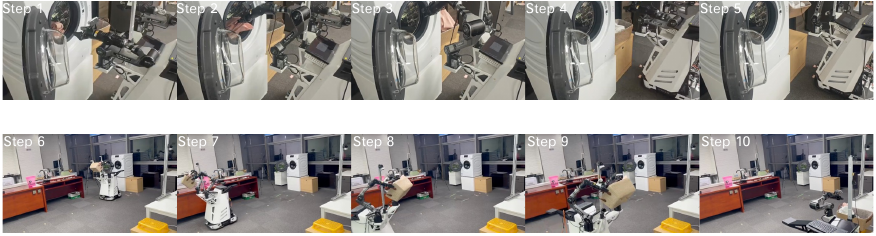}
\caption{Deployment of \textit{GigaBrain-0} on the PiPER arms for real-world \texttt{laundry baskets moving}.}
\label{fig:baskets_demo}
\end{figure}

\subsection{Downstream Task Results}
The most effective validation of \textit{GigaWorld-0} as a data engine for embodied AI lies in its ability to enhance the training of VLA models. To this end, we train {GigaBrain-0}~\citep{gigabrain} on data synthesized by \textit{GigaWorld-0}. The resulting policies demonstrate strong real-world performance and significantly improved generalization across diverse robotic tasks. {GigaBrain-0} succeeds in dexterous manipulation tasks such as \texttt{Laundry Folding} (Fig.~\ref{fig:laundry_demo}) and \texttt{Paper Towel Preparation} (Fig.~\ref{fig:papertowel_demo}), long-horizon mobile manipulation tasks including \texttt{Juice Preparation} (Fig.~\ref{fig:juice_demo}) and \texttt{Table Bussing} (Fig.~\ref{fig:table_demo}), as well as dynamic mobile operations like \texttt{Boxes Moving} (Fig.~\ref{fig:boxes_demo}) and \texttt{Laundry Baskets Moving} (Fig.~\ref{fig:baskets_demo}). These results highlight the fidelity, diversity, and task coverage of the synthetic data generated by \textit{GigaWorld-0}, enabling robust policy learning without extensive real-world demonstrations.
A comprehensive quantitative analysis of task success rates, robustness, and ablation studies can be found in GigaBrain-0~\citep{gigabrain}.

\section{Conclusion}

In this work, we presented \textit{GigaWorld-0}, a scalable and controllable world model designed to serve as a high-fidelity data engine for embodied AI. By unifying photorealistic video generation with geometrically consistent and physically grounded 3D scene simulation, \textit{GigaWorld-0} enables the efficient synthesis of diverse, instruction-conditioned interaction data spanning novel textures, object configurations, and camera viewpoints, circumventing the cost and scalability bottlenecks of real-world data collection. Through rigorous evaluation across multiple embodied benchmarks, we demonstrated that training VLA policies on \textit{GigaWorld-0}-generated data leads to significant improvements in task success, robustness, and zero-shot generalization in real-world robotic environments.

Looking ahead, our work lays the foundation for several compelling directions in world model research for robotics. First, while \textit{GigaWorld-0} currently functions as a powerful \textit{data engine}, a natural evolution is to deploy it as an interactive \textit{policy environment} for model-based reinforcement learning—enabling agents to safely explore, plan, and refine behaviors in simulation before real-world execution. Second, world models like \textit{GigaWorld-0} may ultimately learn universal priors over physical dynamics, semantic affordances, and task structure, allowing them to transition from passive data generators to active \textit{policy co-designers} that propose plausible action sequences or decompose complex tasks into executable subgoals. Finally, closing the loop between real-world experience and synthetic generation, where robot rollouts continuously improve the world model, which in turn produces higher-quality training data, could enable self-improving robotic systems capable of lifelong, autonomous learning. We hope \textit{GigaWorld-0} accelerates progress toward this vision and inspires broader community exploration of world models as central infrastructure for the next generation of embodied intelligence.

\clearpage
\setcitestyle{numbers}
\bibliographystyle{plainnat}
\bibliography{main}

\end{document}